\documentclass[11pt,a4paper]{article}
\usepackage[hyperref]{acl2019}
\usepackage{times}
\usepackage{latexsym}

\usepackage{url}
\usepackage{dialogue}
\aclfinalcopy 

\title{Towards Negotiative Dialogue for the Talkamatic Dialogue Manager}

\author{Staffan Larsson \\
  CLASP\\Dept. of Philosophy, Linguistics\\ and Theory of Science\\ Gothenburg University \\
  \texttt{sl@ling.gu.se} \\\And
  Alexander Berman \\
  Talkamatic AB \\
  Gothenburg, Sweden\\
  \texttt{alex@talkamatic.se}\\\And
  David Hjelm \\
  Talkamatic AB \\
    Gothenburg, Sweden\\
  \texttt{david@talkamatic.se}}

\date{}

\begin{document}
\maketitle
\begin{abstract}
The paper describes a number of dialogue phenomena associated with negotiative dialogue, as implemented in a development version of the Talkamatic Dialogue Manager (TDM). This implementation is an initial step towards full coverage of general  features of negotiative dialogue in TDM. 
\end{abstract}


 \section{Introduction}
 
 The work described in this paper is part of the 3-month phase 1 of the project Tala\footnote{Tala  (Easy-to-use Development Environment for Advanced Dialogue Systems) is funded by The Swedish Post and Telecom Authority (PTS).}. During phase 1, we have  started development of the Tala SDK, a tool for 3rd party creation, validation and deployment of  Dialogue Domain Descriptions (DDDs), allowing developers to adapt the Talkamatic Dialogue Manager (TDM) to new domains. We have also developed  TDM further, by adding a number of dialogue features related to negotiative dialogue. 
 
 Adding support for these features is an example of how TDM can be extended to offer support for dialogue features going beyond simple form-based dialogue, thus extending the space of applications of conversational AI while eliminating the need for developers to provide domain-specific solutions for general dialogue phenomena. 

\section{The Talkamatic Dialogue Manager}

The Talkamatic Dialogue Manager \cite{LB2016} is a commercial platform for conversational AI. 
TDM is the result of an effort to build a dialogue manager on sound engineering principles, exploiting knowledge from  research about human dialogue. The principles that have guided TDM design include (1) Apply general solutions to general problems, and (2)  Don't mix different kinds of knowledge.
These principles have led to an architecture where knowledge about the domain (e.g., telephony or navigation) is separated from general knowledge about dialogue. This means that app developers can focus on defining domain-specific knowledge, such as semantic ontology and their corresponding natural language forms.

The overall logic governing the dialogue is contained within TDM, while domain-specific knowledge, such  as dependencies between various kinds of information in the form of dialogue plans, are kept in the DDDs. Hence, when an app developer builds a new app, there is no need to extend or modify the dialogue manager as such.


 Some aspects of flexible dialogue flow are present in systems like Siri and DialogFlow \cite{larsson:godial,larsson-2017-user,harms2018approaches}, and some are supported by the VoiceXML standard. However, compared to  other systems TDM offers a   wider and more complete range of flexible dialogue behaviours. 


\section{Data collection and analysis}

After an initial micro-scale data collection and analysis, a number of frequent dialogue phenomena related to negotiative dialogue were noticed that could merit implementation in TDM. Some of these phenomena were selected for implementation in a development version of TDM. 

Concretely, the data collection was done as follows: two people communicated via a text chat, one playing the role of "system", looking up phone numbers on a web page, and the other playing the role of "user", requesting phone numbers for various people. In the analysis phase, we looked in particular for frequent dialogue phenomena that were not straightforwardly covered by the current version of TDM. Below is one of the dialogues collected (translated from Swedish to English). The caller is C and the responder is R.

\begin{dialogue}
  \speak{C} 
  hello I need the number for an Anna Andersson in Gothenburg
 \speak{R}  There are fairly many Anna Andersson in Gothenburg. Do you know anything more about where she lives?

   \speak{C} In Old Masthugget
 \speak{R} Do you know which street it is?

  \speak{C} no sorry
    \speak{C} near Slottsskogen I believe
  
\speak{R} There are two Anna Andersson and two Anne Andersson in that area. Is  Anna the name?

  \speak{C} yes
 
\speak{R} In that case one lives on Sj\"omansgatan 13D and the other on Kompassgatan 16A
\speak{R} Is it one of those do you think?

  \speak{C} yes very possibly. can you see their age?
 
\speak{R}  yes. Anna on Sj\"omansgatan is 42 years and Anna on Kompassgatan is 40 years.

  \speak{C} ah interesting
   \speak{C} I'm pretty sure she is 42

\speak{R} OK. In that case her number is NNN-NNN NN NN


  \speak{C} great, thanks

\speak{R} you're welcome

  \speak{C} Can I get the number for the other one as well? The one who is 40

\speak{R}  ok. her number is NNN-NNN NN NN

  \speak{C} perfect
   \speak{C} thanks!

\end{dialogue}

\section{Dialogue features}

In Tala phase 1, intital steps were taken towards the implementation of negotiative dialogue for TDM. 
 The implemented dialogue features are briefly described below, with examples of dialogues handled by the development version of TDM.

\subsection{Asking about alternatives}

Many dialogue systems can handle form-based dialogue, where information is collected from the user, resulting in a database search whose results are communicated to the user. Some systems (including TDM) can also deal with dialogue where a database search is incrementally refined as the user provides more information, with an incrementally shrinking set of alternatives being communicated to the user at each step. However, very few (if any) dialogue system offer general  support for  the kind exploration and comparison of alternatives that frequently precedes a final user choice in human-human task-oriented dialogue\footnote{An exception is restaurant search in Siri, where the user can ask questions about individual restaurants and also compare them. Given that the feature is not generally available across domains, however, indicates  this is probably implemented in a domain-specific manner for the restaurant domain, rather than in a general, domain-independent way.}. We have previously argued that this as a central feature of negotiative dialogue \cite{larsson:thesis}
. As part of this, the user can ask questions about one or several alternatives. We have implemented limited support for user questions about alternatives. 
Below is an example of an interaction supported by the development version of TDM. (S = system, U =user.)

\begin{dialogue}
        \speak{U} I want the number for Anna Andersson in Gothenburg
    \speak{S} There are three persons matching your description.
    \speak{U} How old are they?
    \speak{S} Anna Andersson on Olivedalsgatan 12 in Gothenburg is 77 years. Anna Andersson on Vasagatan 11 in Gothenburg is 42 years. Anna Andersson on Kompassgatan 10 in Gothenburg is 31 years.
\end{dialogue}

\subsection{Knowledge precondition questions}

A less expected finding was the frequent occurrence of questions of the form "Do you know...?", i.e., acts of asking whether the user knows the answer to a question $Q$ rather than asking $Q$ directly. For example, instead of asking "On what street does the person live?", the responder 
would ask "Do you know on what street the person lives?". We refer to such questions as Knowledge Precondition Questions (KPQs) since they (interpreted literally) are asking whether the dialogue partner has the knowledge necessary to answer the embedded question. This type of question has previously been described informally by e.g. \citet{schegloff1999discourse}, who argues that it can be seen as a type of indirect speech act, and more recently in a dialogue systems setting by \citet{moore2018natural}. 

There appears to be a reasonable rationale for asking KPQs in cases where the asker cannot be sure if the answerer has the desired information, such as in the case of street name in the phone directory domain. (By contrast, the caller can be expected to know at least the name of the person they want to call.) By asking a KPQ instead of a direct wh-question, it also becomes easier for the answerer to communicate that they do not know the answer to the question, by simply saying "no". To do this in response to a wh-question, they would have to say "I don't know" or something to that effect, which can be expected to be a less preferred answer.

Typically, KPQs are answered by providing an answer to the embedded question ("The person lives in Gothenburg") or negatively ("No"). In principle they could also be answered with a "yes", although we have not observed this. In any case, we treat these questions as a hybrid between wh-questions and yes/no-questions. This requires that the system is able to correctly interpret and integrate answers to both the overt yes/no KPQ and the embedded wh-question.

In our first example, the user responds to the KPQ with an answer to the embedded wh-question, resulting in narrowing down the number of alternatives:

\begin{dialogue}
    \speak{U} I need the phone number for Anna Andersson 
    \speak{S} There are 4345 persons matching your description. Do you know the city?
    \speak{U} Gothenburg
    \speak{S} OK, there are 86 persons matching your description. Do you know the street name?
\end{dialogue}

In our second example, the user responds negatively to the KPQ, resulting in the system proceeding to the next question (also a KPQ):

\begin{dialogue}
    \speak{U} I need the phone number for Anna Andersson 
    \speak{S} There are 4345 persons matching your description. Do you know the city?
    \speak{U} No
    \speak{S} OK. Do you know the street name?
\end{dialogue}

Finally, a perhaps less common case is where the user responds positively to the direct KPQ, but not to the embedded wh-question. The system then asks the embedded wh-question directly.

\begin{dialogue}
    \speak{U} I need the phone number for Anna Andersson 
    \speak{S} There are 4345 persons matching your description. Do you know the city? 
    \speak{U} Yes
    \speak{S} OK. What city does the person live in?
        \speak{U} Gothenburg
    \speak{S} OK, there are 86 persons matching your description. Do you know the street name?
\end{dialogue}

\subsection{Modifying search criteria}
 
 As can be seen at the end of the transcribed dialogue in Section 3, callers may wish to revise their search after their initial question has been answered. We have improved the support for this dialogue feature in TDM, as exemplified in the dialogue below (which is a continuation of the dialogue in Section 4.1):
 
 \begin{dialogue}
     \speak{S} Returning to the phone number. There are three persons matching your description. Do you know the street name?
\speak{U} Hm, I think she is 42 years old.
\speak{S} OK. The phone number is NNN-NNN NN NN.
\speak{U} What is the phone number for the one who is 31 years old, just in case I'm wrong?
\speak{S} The number is NNN-NNN NN NN
\end{dialogue}

Note that U chooses not to answer the question about street name, but instead provides the age, which allows the system to narrow down the search to a single person. After receiving a phone number, U modifies the search by supplying another age, and receives the corresponding answer.

\begin{figure*}
    \centering
    {\small
    \begin{verbatim}
 1 <domain name="PhoneDirectoryDomain">
 2   <goal type="perform" action="top">
 3     <plan>
 4       <forget_all/>
 5       <findout type="goal"/>
 6     </plan>
 7   </goal>
 
 8   <goal type="resolve" question_type="wh_question" predicate="phonenumber">
 9     <plan>
10       <findout type="wh_question" predicate="person"/>
11       <invoke_service_query type="wh_question" predicate="phonenumber"/>
12     </plan>
13   </goal>

14   <goal type="resolve" question_type="wh_question" predicate="age" 
15         max_answers="3" alternatives_predicate="person">
16     <plan>
17       <findout type="wh_question" predicate="person"/>
18       <invoke_service_query type="wh_question" predicate="age"/>
19     </plan>
20   </goal>

21   <parameters question_type="wh_question" predicate="person" 
22               source="service" incremental="true">
23     <ask_feature predicate="person_name"/>
24     <ask_feature predicate="person_city" kpq="true"/>
25     <ask_feature predicate="person_street_name" kpq="true"/>
26   </parameters>
27 </domain>
\end{verbatim}}
    \caption{Dialogue plans for phone directory domain}
    \label{fig:my_label}
\end{figure*}

\section{Using general solutions in a domain}

To illustrate what is needed by a dialogue developer needs to do to make use of the dialogue features described above, Figure 1 shows TDM dialogue plans for a proof of concept implementation of negotiative dialogue features in the telephone directory domain. The XML format for TDM dialogue plans is described further in \citet{LB2016} and will not be explained in detail here.  

To render a question as a knowledge precondition question, the attribute {\small\texttt{kpq="true"}} is added as in line 24 (and 25). This results in TDM being able to handle the dialogues in Section 4.2 above\footnote{Provided appropriate natural language generation and interpretation facilities, not described in this paper}.

To enable asking the system about alternatives, the tags {\small\texttt{max$\_$answers}} and {\small\texttt{alternatives$\_$predicate}} are used as in line 15. For the phone directory domain, we implemented the possibility for the user to ask about the age of a person in the directory (lines 14-20). To answer this, the system needs to figure out which person is being referred to (same as when the user is asking about a phone number, lines 8--13). To search for persons, a number of features of the person are collected by the system (lines 21--26). However, the addition of line 15 means that if a set (with an arity of 3 or less) of alternative persons (arguments of the predicate {\small\texttt{person}}) have already been established, the user can also ask how old they are, and get the answer for each of the persons individually as shown in Section 4.1 above.

Allowing for modification of search parameters after completed search does not require any additions to the DDD.  

\section{Future work}
In phase 2 of Tala we plan to implement full support for negotiative dialogue in TDM, and to complete the Tala development tool.

 


\end{document}